%% file: emnlp2020.tex
\documentclass[11pt,a4paper]{article}
\usepackage[hyperref]{emnlp2020}
\usepackage{times}
\usepackage{latexsym}

\usepackage{booktabs}
\usepackage{graphicx}
\usepackage{todonotes}
\usepackage{paralist}
\usepackage{enumitem}

\makeatletter
\g@addto@macro{\UrlBreaks}{\UrlOrds}
\makeatother

\newcommand{\dataset}{\textsc{QABriefDataset}}

\newcommand{\qabriefmodel}{\textsc{QABriefer}}

\usepackage{microtype}

\aclfinalcopy 

\title{Generating Fact Checking Briefs}

\newcommand{\fair}{$^1$}
\newcommand{\loria}{$^3$}
\newcommand{\ucl}{$^2$}
\newcommand{\uclfair}{$^{1,2}$}
\newcommand{\loriafair}{$^{1,3}$}
\newcommand{\cambridge}{$^4$}

\author{Angela Fan\loriafair{},
Aleksandra Piktus\fair{},
Fabio Petroni\fair{},
Guillaume Wenzek\fair{},\\
\bf Marzieh Saeidi\fair{},
Andreas Vlachos\cambridge{},
Antoine Bordes\fair{},
Sebastian Riedel \uclfair{}\\
\\
\fair{}Facebook AI Research \ \ucl{}University College London \ \loria{}LORIA,
\cambridge{}University of Cambridge \\
\{angelafan,piktus,fabiopetroni,guw,marzieh,avlachos,abordes,sriedel\}@fb.com
}

\date{}

\begin{document}
\maketitle

\begin{abstract}
Fact checking at scale is difficult---while the number of active fact checking websites is growing, it remains too small for the needs of the contemporary media ecosystem.
However, despite good intentions, contributions from volunteers are often error-prone, and thus in practice restricted to claim detection.
We investigate how to increase the accuracy and efficiency of fact checking by providing information about the claim before performing the check, in the form of natural language \emph{briefs}. 
We investigate passage-based briefs, containing a relevant passage from Wikipedia, entity-centric ones consisting of Wikipedia pages of mentioned entities, and Question-Answering Briefs, with questions decomposing the claim, and their answers. 
To produce QABriefs, we develop \textsc{QABriefer}, a model that generates a set of questions conditioned on the claim, searches the web for evidence, and generates answers.
To train its components, we introduce \dataset{} 
which we collected via crowdsourcing. 
We show that fact checking with briefs --- in particular QABriefs --- increases the accuracy of crowdworkers by 10\% while slightly decreasing the time taken.
For volunteer (unpaid) fact checkers, QABriefs slightly increase accuracy and reduce the time required by around 20\%. 
\end{abstract}

\section{Introduction}
\input{intro}

\begin{figure*}[t]
    \centering
    \includegraphics[width=\linewidth]{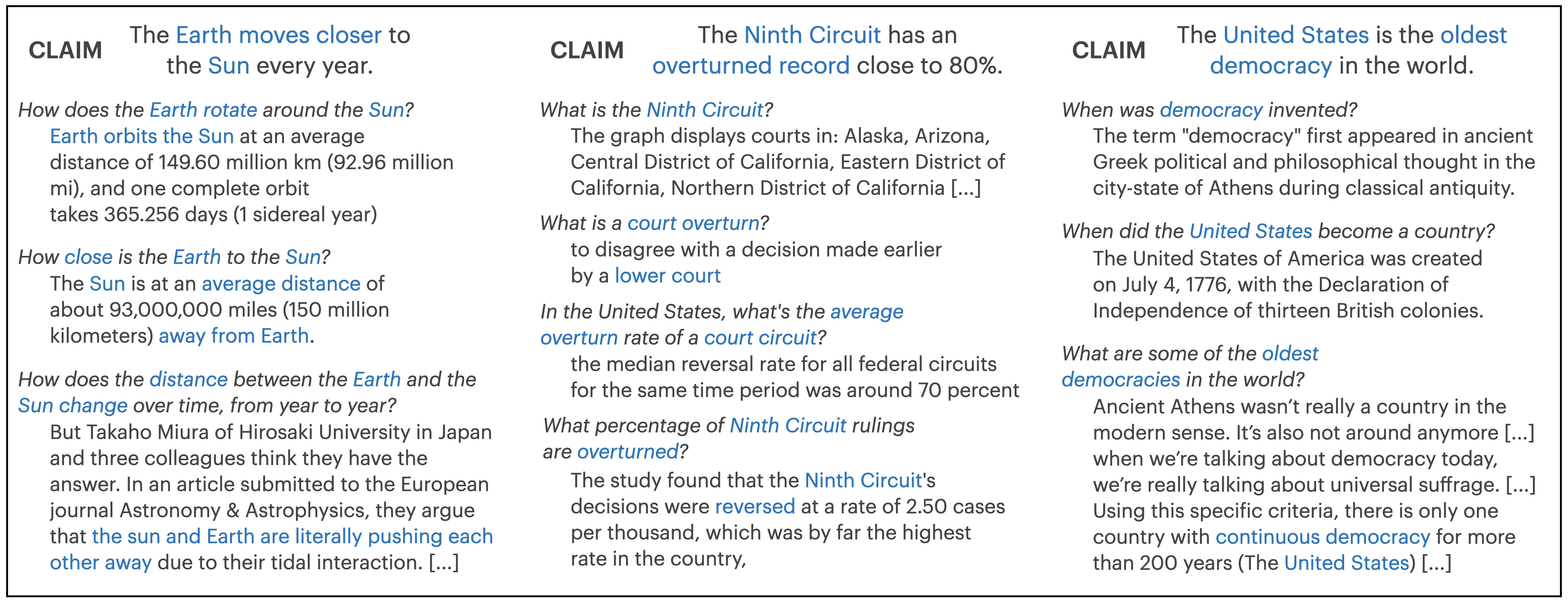}
    \caption{\textbf{Examples of QABriefs in \dataset{}}}
    \label{fig:many_examples}
\end{figure*}

\section{Briefs for Fact Checking}

Fact checkers must comprehend each part of a claim, which requires gathering information about a wide range of concepts--- a precise definition of a term, how a politician voted, or the exact contents of a bill. 
Such knowledge is available in many sources: knowledge bases, statistical reports, or on the internet.
We introduce the notion of \emph{briefs} to provide relevant information to fact checkers---as if \emph{briefing} them before fact checking---
and explore three possible forms: Passage Briefs, Entity Briefs, and Question Answering Briefs.
We show how they can be constructed with modern NLP approaches. 

\subsection{Passage Briefs}

To provide information before checking a claim, Passage Briefs consist of relevant passages retrieved from Wikipedia.
For the claim in Figure~\ref{fig:example_briefs}, information about the history and implementation of social security in the United States is retrieved and presented as background for the fact checker.
To generate Passage Briefs, we identify relevant Wikipedia passages for each claim. Based on the 
results by~\citet{lewis2020retrieval} on open-Wikipedia tasks, we use the Dense Passage Retriever (DPR)~\cite{DBLP:journals/corr/abs-2004-04906}. This state of the art, pretrained retriever model 
learns a representation of questions and possible relevant paragraphs. In our case, we provide the claim as input instead of the question, rank the outputs, and select the top ranked passage. We limit to 500 tokens for readability.
Initial experiments suggested web-based Passage Briefs returned poor results for most claims, as it relied on a finding a single passage addressing the entire claim, so we keep the Passage Brief focused on Wikipedia. Further, DPR is trained on Wikipedia, and we found the best performance within this domain.

\subsection{Entity Briefs}

Passage briefs provide information from a single passage, but claims are complex and often require multiple pieces of information from different sources. Thus we propose entity briefs that focus on each entity referenced in the claim. 

Entities in each claim are identified with \textsc{blink}~\cite{wu2019zero}, a model trained on Wikipedia data that links each entity to its nearest Wikipedia page. \textsc{blink} combines a bi-encoder~\cite{urbanek2019learning,humeau2019poly} that identifies candidates with a cross-encoder that models the interaction between mention context and entity descriptions. For each entity, we retrieve its Wikipedia and provide the first paragraph in the brief. In  Figure~\ref{fig:example_briefs}, \emph{Franklin Roosevelt} is an entity, and the brief communicates he is \emph{an American politician who served as the 32nd president of the United States [...]}. 
However, unlike Passage Briefs, if several entities are identified, information from multiple pages is displayed in an Entity Brief.

\subsection{Question Answering Briefs}

Entity briefs provide information about entities mentioned in the claim, but not necessarily the evidence needed for the claim in question. For this reason we propose 
QABriefs, which decompose fact checking 
into a set of questions and answers. E.g.\ the claim in Figure~\ref{fig:example_briefs} could be split into understanding what social security is, identifying who invented the concept, and finally where Franklin Roosevelt got the idea.  
Each step can be written into a question --- \emph{What is social security? Who invented social security?} --- that is then answered. The decomposition into question-answer pairs is likely to be better amenable to the current generation of information retrieval systems, which typically assume simpler information needs, e.g.\ most QA datasets have questions about single factoids.
Unfortunately, there are no existing datasets or models available to create QABriefs. Next, we describe how we create a dataset (Section~\ref{sec:dataset}) and a model (Section~\ref{sec:model}) to produce QABriefs.

\section{QABrief Dataset}
\label{sec:dataset}

To train and evaluate models to generate QABriefs, we collect a dataset of questions based on claims, together with answers to those questions found on the open web. Crucially, annotators first read the 
article from a fact checking website that describes how the claim was checked, and then decompose the process into questions, for which answers are provided. 
The claims for the dataset are sourced from existing fact checking datasets, specifically \textsc{datacommons}\footnote{\url{https://datacommons.org/factcheck}} and \textsc{multifc}~\cite{augenstein-etal-2019-multifc}. The annotator instructions are in the Appendix and examples are shown in Figure~\ref{fig:many_examples}.

While numerous question generation and answering datasets exist, none of them focuses on using questions and answers to combat misinformation. \dataset{} focuses on this real world problem, with each question grounded in a claim that was actually fact checked. Further, existing datasets are quite different from our usecase --- for example, many datasets are based on Wikipedia, but fact checkers find evidence from other sources. Many datasets have short answer spans, but our questions are complex, so require longer answers. 

\subsection{Question Generation}

Crowdworkers are asked to read a \emph{claim} and its corresponding 
\emph{fact checking article}\footnote{For our running example, the reference article is: \url{https://www.politifact.com/factchecks/2016/dec/16/russ-feingold/was-social-security-basically-invented-university-/}}, which details the investigative process used to perform the fact check. 
After reading the article, crowdworkers write \emph{questions} to reconstruct the process taken by professional fact checkers. For each claim, crowdworkers write two to five questions that are at least five words long and standalone. For instance, the question \emph{why did he do that} is invalid, as it is not clear what \emph{he} or \emph{that} is. We discourage questions with yes/no answers and discourage questions about the same claim from overlapping more than five words.

After the questions are collected, a \emph{question validation} phase is conducted. A separate group of crowdworkers reviews the quality of the questions and flags those
that are redundant and/or otherwise poor quality. For example, questions such as \emph{What evidence is there that [claim] is true?} are rejected. Other instances of questions rejected at this phase include nonsensical questions and questions that simply rephrase the claim. Any questions that do not pass this review are re-annotated.
Finally, a \emph{question clarity} phase is conducted --- crowdworkers read the questions and edit those that are unclear or underspecified. For example, questions may need to have a year added to them to accurately verify a statistic.
Further, additional questions can be added if crowdworkers feel the existing questions are not sufficient. 
This can lead to more than five questions per claim. Spelling errors are highlighted and crowdworkers are encouraged to correct them. 

\subsection{Question Answering}

After each claim is annotated with multiple questions,
we proceed to collect the answers to them.
To answer questions, crowdworkers are given the claim; the source of the claim (for example, the entity who said the quote being checked); and the question. 
Crowdworkers enter a \emph{query} into a \emph{search engine} to find information on the web. The search is restricted from accessing fact checking domains, to prevent the answer from being trivially found on a fact checker's website. The query does not need to be identical to the question, and is often rephrased to find better search results. After reading the returned results, crowdworkers can provide one of three possible answer types:
\begin{itemize}[noitemsep,topsep=0pt]
    \item \emph{Extractive} --- the encouraged option, crowdworkers copy paste up to 250 words as an answer. We focus on extractive answers, as identifying such an answer is more straightforward compared to writing an answer.
    \item \emph{Abstractive} --- if the answer is present in an image or graph, crowdworkers write an abstractive answer of at least 20 words.
    \item \emph{No Answer} --- if no answer can be found, crowdworkers write an explanation of at least 20 words to describe why there is no answer.
\end{itemize}

Next, \emph{validation} is conducted. The questions are complex, so we do not assume the answer is known. Crowdworkers instead flag answers that seem incorrect. For example, if the answer to \emph{How many people live in California} is \emph{three billion}, this would be flagged and re-annotated. A last step is conducted for answers that are \emph{No Answer}. To verify that answers cannot be found, a second group of crowdworkers tries to find an answer. If an answer is found, the \emph{No Answer} annotation is discarded. 

\begin{table}
\centering
\begin{tabular}{l l r}
\toprule
\textbf{Train} & Number of Claims & 5,897 \\ 
& Number of QA Pairs & 18,281 \\ 
\textbf{Valid} & Number of Claims & 500 \\ 
& Number of QA Pairs & 1,431 \\ 
\textbf{Test} & Number of Claims & 500 \\ 
& Number of QA Pairs & 1,456 \\ 
\midrule 
\multicolumn{2}{l}{Avg Number Questions/Claim} & 3.16 \\ 
\multicolumn{2}{l}{Avg Number Words in Questions} & 10.54 \\ 
\multicolumn{2}{l}{Avg Number Words in Answers}  & 43.56 \\ 
\bottomrule
\end{tabular}
\caption{\textbf{Statistics of \dataset{}}}
\label{table:data_statistics}
\end{table}

\begin{figure}[t]
    \centering
    \includegraphics[width=\linewidth]{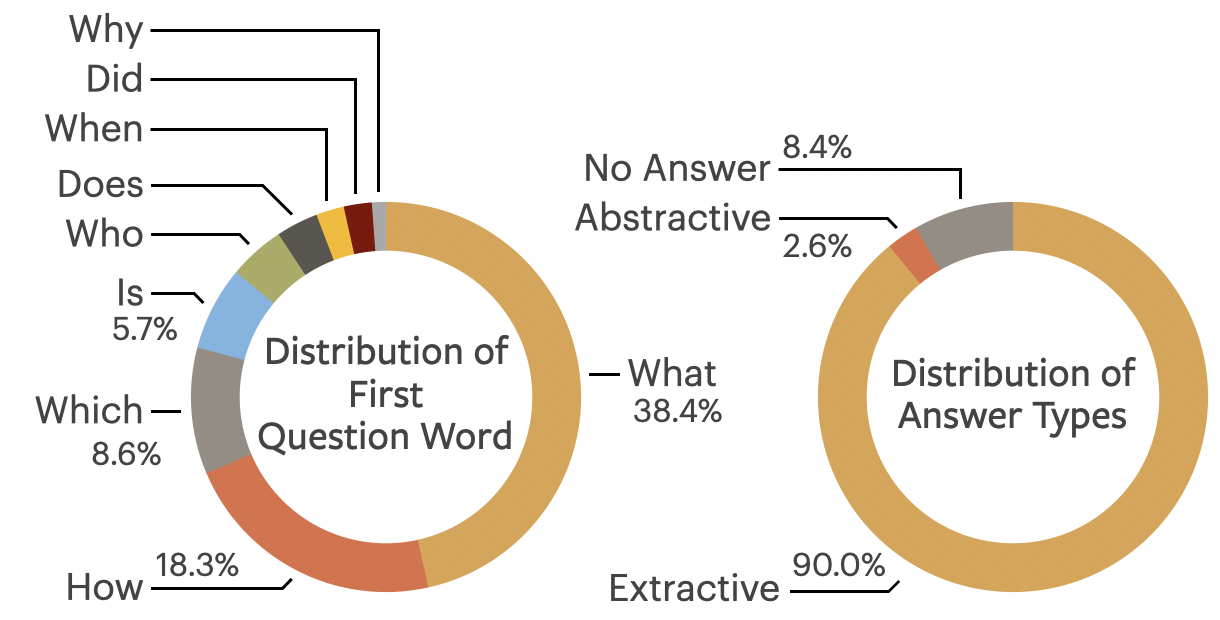}
    \caption{\textbf{Question and Answer Types}}
    \label{fig:distribution}
\end{figure}

\begin{figure*}[t]
    \centering
    \includegraphics[width=\linewidth]{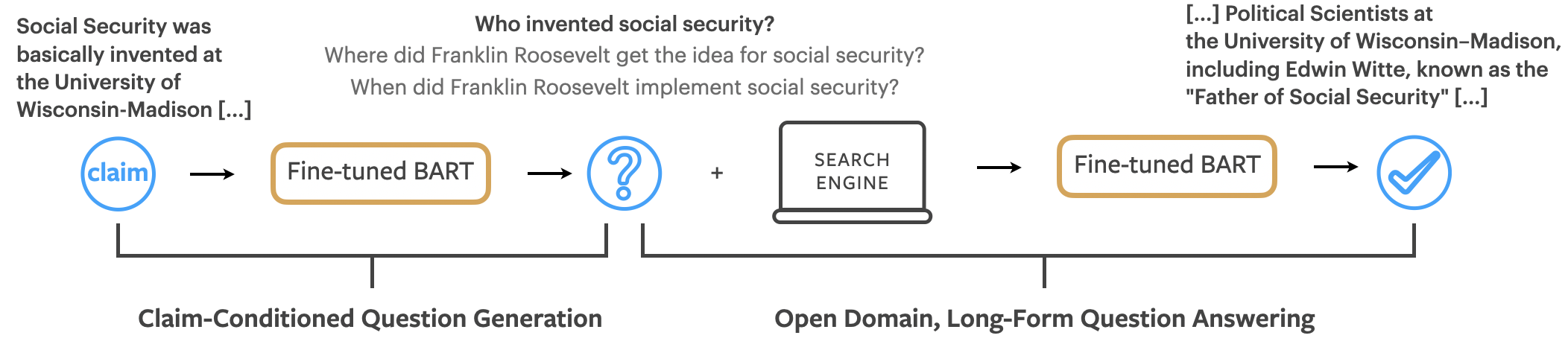}
    \caption{\textbf{QABriefer Model.} First, BART is finetuned to conduct  claim-conditioned question generation and generates a sequence of questions that decompose a fact check. Second, we use an information retrieval system and a second finetuned BART model to extract long-form answers to each question.}
    \label{fig:model}
\end{figure*}

\subsection{QABrief Dataset Statistics}

In summary, \dataset{} includes 6,897 claims and 21,168 questions paired with their answers. We use 500 claims as a validation set and 500 claims as a test set. The validation and test sets include around 1400 questions and answers each. 

We examine the types of questions to analyze the diversity. Table~\ref{table:data_statistics} shows that each claim on average requires around 3 questions to cover the different parts of the claim, and questions contain 10.5 words on average. The questions are quite diverse, as seen in Figure~\ref{fig:distribution} (left), though the majority begin with \emph{What, How, Which} question words. There are few \emph{Why} questions, indicating a focus on verifying factual information, rather than causality.

The answers obtained have mainly extractive annotations, though a small portion of abstractive and no answer options exist (see Figure~\ref{fig:distribution}, right). Answers are around 43.5 words long (Table~\ref{table:data_statistics}), though abstractive answers are generally shorter as crowdworkers must fully write them.

We examined a subset of 50 claims where we conducted multiple data collection trials with the same claim to understand the agreement rate between workers. We found that for the question annotation step, about half of the questions provided by different people on the same claim were very similar and could be considered paraphrases. For example, the questions \textit{Who invented social security} and \textit{Who was the invetor of social security}. For the answer annotation step, the identified answers varied  in length but were often paraphrases --- some crowdworkers tended to select only the specific span that answered the question (e.g.\ an entity name), while others chose several sentences to capture the context.

\section{QABrief Model}
\label{sec:model}

The automatic generation of QABriefs presents numerous modeling challenges. Generating such a brief is a hierarchical process: writing the questions, and then conditioned upon the questions, searching the web and writing the answers. While many question answering datasets exist, questions in \dataset{} are grounded on real claims that were fact checked. 
The diversity of the claims renders reusing questions across claims  unlikely to work, thus precluding the use of retrieve-and-rank approaches \citep{rao-daume-iii-2018-learning}.
Unlike previous question generation models~\cite{du2017learning,duan2017question,tang2017question,zhao2018paragraph} that generate  based on an answer, we treat question generation closer to structured planning --- laying out the format for the entire brief. 

In contrast to most question answering datasets, the length of the answers in \dataset{} are long-form~\cite{fan2019eli5}. For example, the average answer in SQuAD~\cite{rajpurkar2016squad} is four words long, while the average answer in \dataset{} is forty. Further, datasets such as SQuAD, Natural Questions~\cite{kwiatkowski2019natural}, and HotpotQA~\cite{yang2018hotpotqa} are built from Wikipedia, while QABriefs uses the web. 

In this section, we describe \qabriefmodel{} (see Figure~\ref{fig:model}). 
For each claim, the question generation model is used to generate multiple questions. For each question, an evidence document is retrieved using a search engine. 
We take the top search hit as the evidence and retrieve the text from CommonCrawl\footnote{\url{http://commoncrawl.org/}}. Finally, the generated question and retrieved evidence document is provided to the question answering model to generate an answer.

\begin{figure*}[t]
    \centering
    \includegraphics[width=\linewidth]{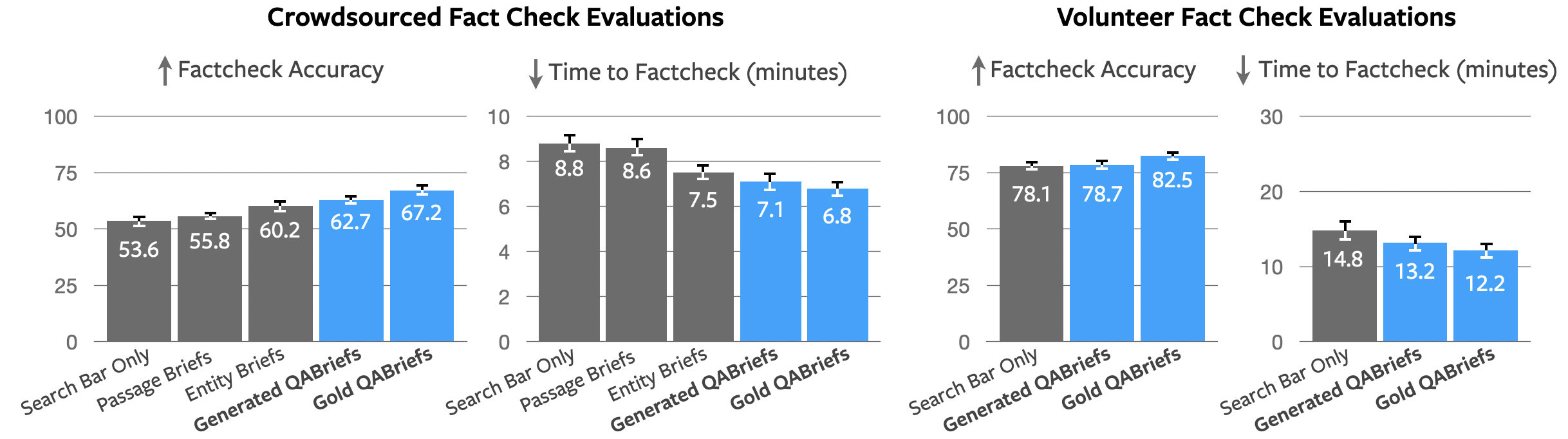}
    \caption{\textbf{Accuracy and Time Taken to fact check} by Crowdworkers (left) and Volunteer fact checkers (right). Briefs of various forms, but particularly QABriefs, increase fact checking accuracy and efficiency.}
    \label{fig:human_eval_results}
\end{figure*}

\begin{figure}[t]
    \centering
    \includegraphics[width=0.9\linewidth]{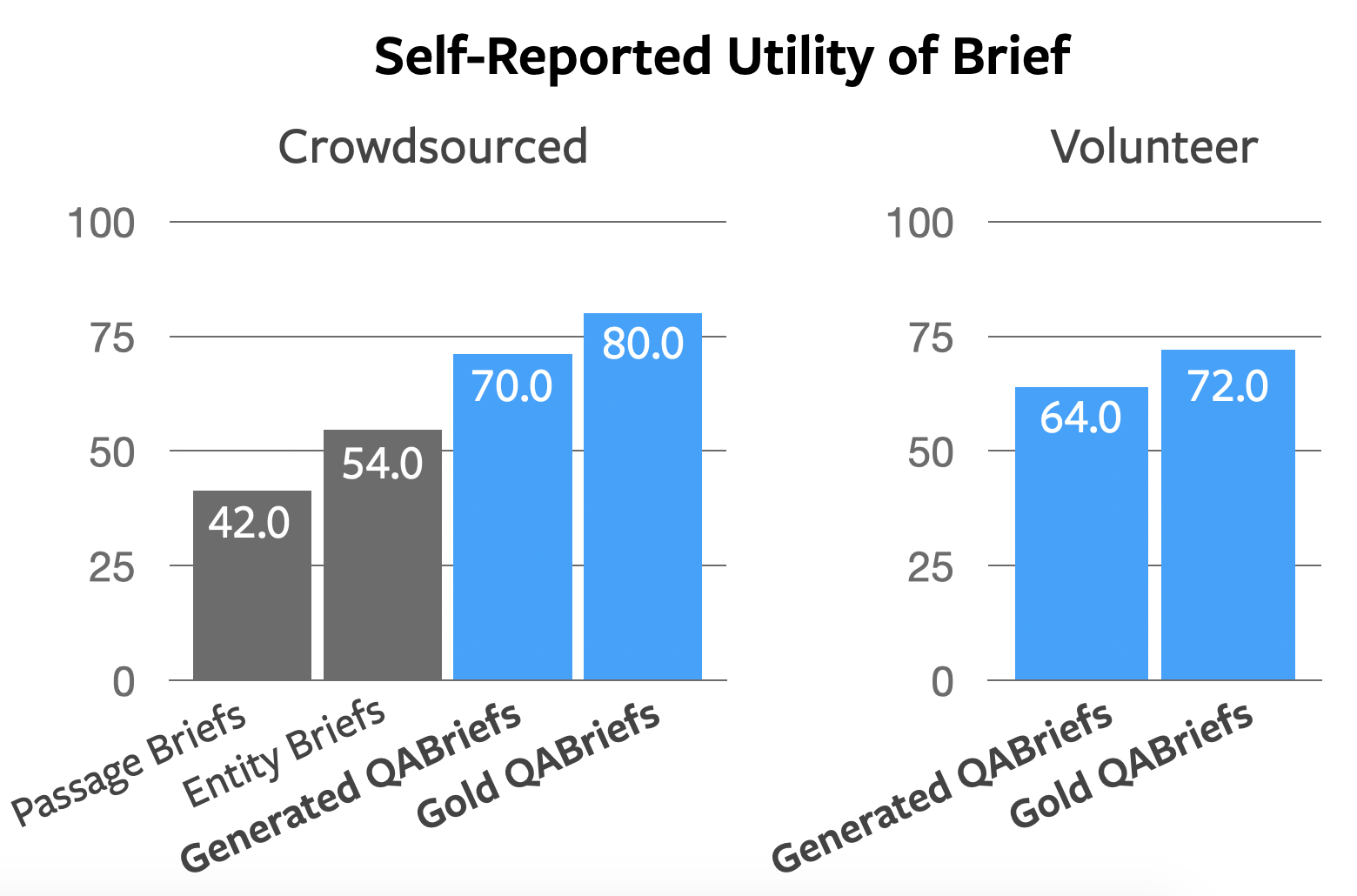}
    \caption{\textbf{Usefulness of Briefs} reported by Crowdsourced and Volunteer Fact Checkers.}
    \label{fig:pref_qa_brief}
\end{figure}

\subsection{Question Generation}

The first step of \qabriefmodel{} is to create the questions that will form the structure of the brief. To create models that can take a claim as input and generate a sequence of questions as output, we use sequence-to-sequence~\cite{sutskever2014sequence} models. 
As \dataset{} is not large enough to train the language model needed for question generation, we leverage advances in pretraining and use \dataset{} to adapt it to the task at hand.
We use BART~\cite{lewis2019bart}, a denoising autoencoder that uses various noise functions and trains to recreate the input. 
In adapting BART for question generation based on claims, 
we explore three options: generating all questions based only on the claim, generating all questions based on the claim and the source of the claim (usually an entity), and generating questions one at a time. To write questions one at a time, the model conditions on the previous questions as well as the claim and source, and needs to predict the subsequent question or an \emph{end of questions} token. 

\subsection{Question Answering}

Given the question-based structure for QABriefs, the second part of the hierarchical process is to identify answers. Models take as input the question and evidence document that annotators indicated to contain the answer, and produce an answer. 
As \dataset{} does not have enough data to train a question answering model from scratch, we use BART finetuned on Natural Questions. 
and subsequently finetune it further on \dataset{}. As the dataset contains extractive and abstractive answers as well as questions where the model must provide an explanation to justify no answer, we use an abstractive approach with a generative model;  abstractive models have shown strong performance on various question answering tasks \cite{lewis2018generative,dong2019unified,radford2019language,raffel2019exploring,lewis2020retrieval}.

\section{Experimental Setup}

Our main question is whether briefs can increase the accuracy and efficiency of fact checking. We focus on human evaluation with both  crowdworkers and volunteers fact checking claims. 

\subsection{Human Evaluation}

\paragraph{Metrics} We evaluate the \emph{accuracy} of a fact check by comparing the verdict from our human evaluators with professionals. The professional fact checking labels are obtained from the \textsc{datacommons} dataset. We measure the \emph{time} taken to fact check from when the task is loaded to when the verdict and explanation is submitted. 

\paragraph{Crowdsourced Evaluators} Crowdworkers on Mechanical Turk are presented with the 500 test set claims and instructed to use a search bar to decide if the claim is true, false, or in the middle. They then write at least 20 words justifying their verdict. We indicate that if a claim is \emph{mostly true} it should be labeled as true, and \emph{mostly false} should be false. 
We discourage the middle option and suggest it should be used only if a verdict cannot be made, to prevent it from being the default. 
Previous work has shown that fine-grained labels, such as \emph{sometimes true}, \emph{half true}, \emph{mostly true} are difficult to calibrate even with professional fact checkers~\cite{lim2018checking}, so we opt for a more 
simpler
scale. 
The search bar queries the open web, but is restricted from searching known fact checking domains. Evaluators either use only the search bar, or are provided with a brief to read before the fact check. 
The same claims are evaluated with all methods. We repeat the study  three times to assess variance. 

\paragraph{Volunteer Evaluators} Crowdsourced evaluation is scalable, but crowdworkers may be less motivated to spend a large amount of time fact checking. Thus, we conduct a smaller scale study using graduate student volunteer evaluators, recruited by asking for those interested in the challenge of fact checking real claims themselves. Volunteers are presented with 100 claims rather than 500, but otherwise conduct the same task as crowdworkers.  Volunteers compare the search-bar-only fact checking process with generated QABriefs and gold QABriefs. We do not evaluate Passage Briefs or Entity Briefs, as we found volunteer fact checking to be less scalable than crowdsourcing.

\subsection{Automatic Evaluation of Model Quality}

To evaluate the quality of question generation, following existing work~\cite{duan2017question}, we use BLEU. To evaluate the quality of question answering, we use F1 score~\cite{rajpurkar2016squad}. 

\subsection{Model Details} 

We use \texttt{fairseq-py}~\cite{ott2019fairseq} to train the \qabriefmodel{}. We use the open-sourced BART model~\cite{lewis2019bart} and suggested finetuning hyperparameters, training for 10 epochs and taking the best epoch by validation loss. To generate, we use beam search with beam size 5. We tune the length penalty to decode such that written questions and answers approximately match the average length in the validation split. Exact training and generation commands, with further experimental details, can be found in the appendix.

\begin{table}
\centering\small
\begin{tabular}{ll}
\toprule
\bf Model & \bf BLEU  \\  
\midrule 
Claim $\Rightarrow$ Qs & 12.8  \\ 
Claim + Source $\Rightarrow$ Qs & 13.2 \\ 
Claim + Source + Prev Qs $\Rightarrow$ Next Q & 13.4 \\ 
\bottomrule
\end{tabular}
\caption{\label{table:question_results} \textbf{Question Generation Models}}
\end{table}

\begin{table}
\centering\small
\begin{tabular}{lll}
\toprule
\bf Model & \bf F1 \\
\midrule 
BART FT on \dataset{} & 30.5 \\ 
BART FT on NQ + \dataset{} & 32.8 \\ 
\bottomrule
\end{tabular}
\caption{\label{table:answer_results} \textbf{Question Answering Models.}}
\end{table}

\section{Results}

We show in human evaluations that fact checking efficiency and accuracy are improved with briefs.

\subsection{Briefs Increase Fact Checking Quality}

We examine the accuracy of crowdsourced and volunteer fact checkers when presented ---in addition to a search bar--- with different types of briefs: Passage, Entity, and QABriefs. For QABriefs, we examine briefs generated by \qabriefmodel{} and the \emph{Gold} briefs annotated in \dataset{}. We 
compare briefs 
against
a \emph{search bar only} baseline. 

As shown in Figure~\ref{fig:human_eval_results} (left), when crowdworkers are presented with briefs, fact checking accuracy  increases, even when taking into account variance in three repeated trials. The Passage Briefs are not more helpful in terms of accuracy compared to using the search bar alone, but Entity Briefs and QABriefs are both better than this baseline. Providing Gold rather than generated QABriefs performs best --- suggesting modeling improvements could help bridge the gap. For crowdworkers, using briefs slightly reduces the time taken (from 8.8 minutes on average to around 7), but the overall time spent is low compared to professionals, who spend from 15 minutes to one day~\cite{hassan15quest}.

For volunteer fact checkers (Figure~\ref{fig:human_eval_results}, right), accuracy across all methods is higher compared to crowdworkers. Providing the Gold QABrief remains the best, though the gap is smaller than for crowdworkers. Providing the QABrief slightly decreases time taken to fact check. Note that the average volunteer spends twice the amount of time compared to a crowdworker, and this thoroughness probably contributes to higher accuracy, as well as the smaller improvement from providing briefs.

\subsection{QABriefs are Preferred}

Next, we further contrast QABriefs with Passage and Entity Briefs. We ask evaluators to consider if the brief made the fact check \emph{easier} or provided useful \emph{background context}.
Crowdworkers rated QABriefs helpful twice as often as Passage Briefs (In Figure~\ref{fig:pref_qa_brief}). 
When evaluators submit a fact check, they must write an explanation for their reasoning. Qualitatively examining these, we found many references to the QABrief. Evaluators noted that \emph{based on [the QABrief], I searched for [X evidence]}. We hypothesize that the question-answer format may be easier to read, as it is naturally organized and possibly less redundant.

\subsection{Generating QABriefs with \qabriefmodel{}} 

Lastly, we assess the performance of our proposed \qabriefmodel{} model. We display the BLEU scores for our proposed Question Generation models in Table~\ref{table:question_results} and find that iteratively writing questions one by one is the best performing method. Further, providing information about the source of the claim (usually the entity who made the claim) provides better results. Question Answering results are shown in Table~\ref{table:answer_results}. We find that first fine-tuning on a large question answering dataset, Natural Questions (NQ), and further fine-tuning on \dataset{} provides the best results. Likely, this is because BART is a general purpose generative model, so fine-tuning for question answering first on a much larger dataset is useful.

\section{Related Work}

Previous work in NLP has focused on claim veracity. It has been treated as a classification problem~\cite{wang-2017-liar}, often using stance detection~\cite{riedel2017simple}. The FEVER Challenge~\cite{thorne-etal-2018-fever} proposed providing provenance for a decision along with classification, and various approaches developed combine information retrieval with stance detection or question answering~\cite{li-etal-2018-end,lee-etal-2018-improving}.
Question generation and answering has been considered in the context of FEVER \cite{jobanputra-2019-unsupervised} --- 
the focus was on eliciting the right answer from a question answering system rather than improving the accuracy and efficiency of human fact checkers.

However, FEVER is based on modified Wikipedia sentences, not real world claims, which are arguably more difficult. 
To address this
\citet{hanselowski-etal-2019-richly} considered the claims fact checked by the website Snopes, but used the reports accompanying them as evidence instead of finding the evidence directly. \citet{popat-etal-2018-declare} and \citet{augenstein-etal-2019-multifc} used search engines, but without ensuring that they provide evidence supporting/refuting the claim instead of being related to it or that they were not fact checking reports. Finally, \citet{kochkina-etal-2018-one} used responses on social media for rumour verification, but did not address evidence finding.

Various work studies how to improve the fact checking process. Analysis shows accuracy can improve by providing feedback~\cite{hill2017learning}, additional time~\cite{bago2020fake}, tooling~\cite{karduni2019vulnerable}, or training~\cite{zhang2018structured}. These works are complementary to ours --- we provide support in the form of briefs. Studies emphasize that current solutions for fully automated fact checking face various challenges~\cite{graves2018understanding} that must be addressed with interdisciplinary research~\cite{karduni2019human}. Developing tools to aid human-in-the-loop fact checking has received increasing attention, from NLP to human-computer interaction and psychology, often with positive results when tested with journalists~\cite{miranda2019automated} and professionals~\cite{lurie2019challenges}.

\section{Discussion}

While our experiments show a generally positive impact of briefs for human fact checking, it is important to put them into a broader  perspective.

\paragraph{Briefs for Professional Fact Checkers}

Crowdworkers and professional fact checkers perform different tasks under very different circumstances. Professionals often investigate alternative interpretations and produce an explanation of their process in an article. They often have years of experience and must check a variety of claims. Consequently, we do not claim that briefs will make a difference in their work. 
Nevertheless, QABriefs can provide insights into the fact checking process. As the QABrief dataset was created using professional fact checking articles describing how a claim was checked, by decomposing a claim into multiple components, we can encourage a more structured fact checking process.

\begin{figure}[t]
    \centering
    \includegraphics[width=0.9\linewidth]{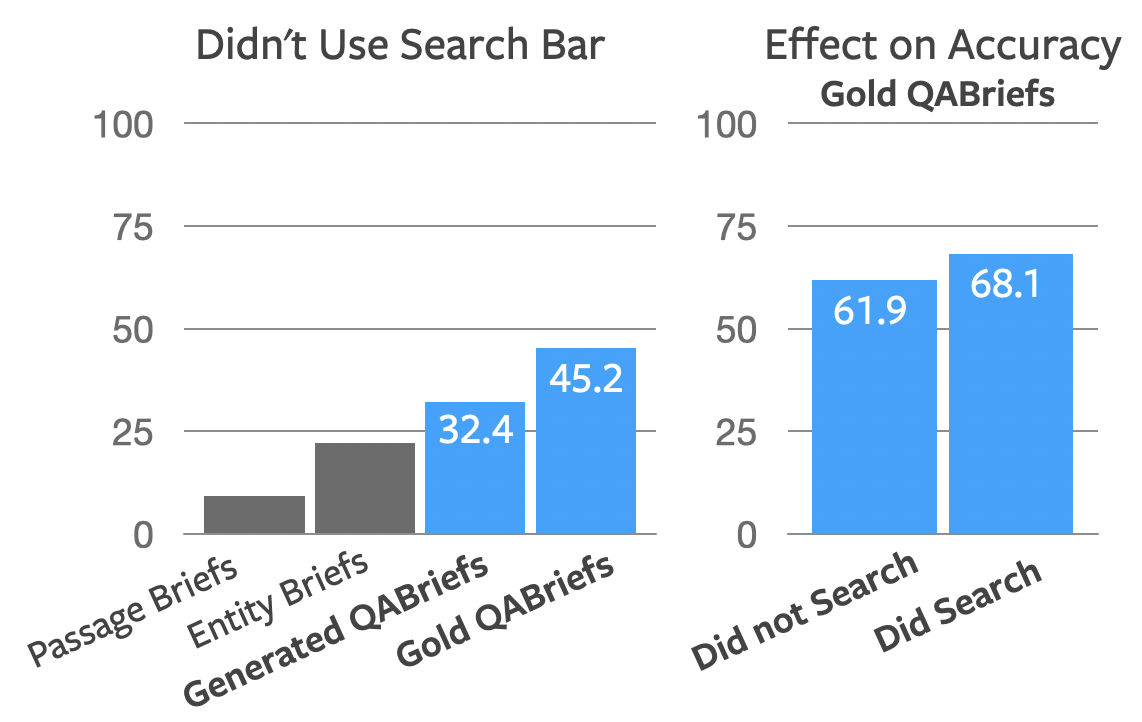}
    \caption{\textbf{Overconfidence} when given a QABrief.}
    \label{fig:use_search_bar}
\end{figure}

\paragraph{Biases introduced by Briefs}

While briefs can increase accuracy, they can introduce biases.
We found that providing a QABrief increased confidence --- many submitted their fact check based on the brief alone, without the search bar. 
Figure~\ref{fig:use_search_bar} (left) displays that around 45\% of crowdworkers did not use the search bar when given the Gold QABrief, even though accuracy without the search bar is reduced.
Briefs aid accuracy and efficiency, but are not fully sufficient to produce a verdict.

\paragraph{Metrics for Factchecking}

We focus on improving fact checking accuracy, but we note that agreement amongst professionals is not 100\%~\cite{lim2018checking}.
Professionals often agree if part of a claim is true or false, but disagree on the importance~\cite{lim2018checking} or pursue different directions for checking the claim~\cite{marietta2015fact,amazeen2016checking}. Different fact checkers have different scales, which are not calibrated. Nevertheless, improving the accuracy of crowd sourced fact checkers is still reflective of agreement with professionals.

\section{Conclusion}
We propose the concept of fact checking briefs, to be read before performing a fact check. Crucially, we develop \qabriefmodel{} and release the accompanying \dataset{}, to create QABriefs. We show in extensive empirical studies with crowdworkers and volunteers that QABriefs can improve accuracy and efficiency of fact checking.

\bibliography{anthology,emnlp2020}
\bibliographystyle{acl_natbib}

\clearpage 
\section{Appendix}

\subsection{Dataset Analysis}

In this section, we describe qualitative observations about \dataset{} to provide more insight. 

\paragraph{How are fact checks decomposed into questions?}

We analyze the strategies taken by annotators to decompose the fact checking process of a claim into multiple questions. There are several distinct strategies that emerge:

For questions about comparison, annotators usually write 1-2 questions validating the first part of the comparison and 1-2 questions validating the second part of the comparison.

For questions about historical events, annotators usually clarify the entities involved and clarify the background. Annotators often ask questions about time and location. Several questions of the form \emph{Did X event really happen} arise, but are often filtered by later steps of the dataset collection process (see description later in this Appendix). 

For questions about what an individual may have said, annotators adopt a strategy very similar to professional fact checkers. A common trend in misinformation is misattribution, or saying an individual said a statement when they did not. Often, a misalingment in time or location can reveal this --- if the person was not yet born, for example. Annotators often ask many questions to try to uncover this.

\paragraph{How are annotators finding answers?}

In many standard question answering datasets, the question-answer pairs already exist. For example, in TriviaQA~\cite{joshi2017triviaqa}, the questions and answers are from Trivia enthusiasts, and in ELI5~\cite{fan2019eli5}, the questions and answers are from Reddit question answering subreddits. Other datsets collect questions and answers, but focus on identifying extractive answers in Wikipedia, an arguably easier task than finding them on the web. In SQuAD~\cite{rajpurkar2016squad}, questions are often written by modifying a sentence of Wikipedia into a question. In Natural Questions~\cite{kwiatkowski2019natural} and MSMarco~\cite{nguyen2016ms}, the questions are real questions submitted to Google and Bing search engines, but the answers are much more straightforward (short, extractive spans). 

In contrast, \dataset{} faces challenges because the questions are complex and the answers must be found on the open web. In initial experiments, we attempted to restrict only to Wikipedia, but found that a large quantity of the questions were annotated with \emph{No Answer}. To find answers on the web is a difficult task, as many answers depend heavily on context. Checking statistics, for example, is particularly difficult, as the year must be correct. We focus on using automated checks, described later on in this Appendix, to check for high quality answers. Further, we spot checked answers manually for quality control. 

We analyzed the main strategies taken to find answers. About 50\% of the annotators directly enter the question in the search bar, but the other 50\% mainly use keyword searches to find better results. Around 83\% of annotators only use the search bar once, but the rest use the search bar two to four times to refine their search query. Note this search query data will be released as part of \dataset{} as well. 

Most annotators submit an answer from the first three search results. Unfortunately, our interface cannot capture how many search results they opened and read before submitting a response. If Wikipedia was in the top search result, most annotators tended to submit a response from Wikipedia.

\subsection{Additional Human Evaluation Results}

In this section, we present additional results from our human evaluation studies. We contrast the process taken by professionals with our volunteer evaluators, analyze if evaluators can accurately assess how difficult a claim is to fact check, and display more detailed results to examine the time taken to fact check a claim. 

\paragraph{Fact Checking Process of Non-Professionals}
In contrast to professionals, we find that crowdworkers and volunteer fact checkers often act on more general understanding rather than validating every detail. 
For example, for some claims, explanations written for a verdict included \emph{It's not possible because the government cannot enforce}, but no evidence is cited.
Over-reliance on common sense can lead to less evidence-based decision-making, and most likely contributes to less time-intensive checks compared to professionals. 
Another instance that commonly arises is checking certain statistics, such as \emph{how many people purchased X item}. A professional fact checker will cross-reference the year carefully, examine how purchases are quantified in stores and through online retailers, and break it down by country. A volunteer examining the same claim will investigate with a search engine, but likely trust a holistic number they find, rather than breaking it down. 

\paragraph{Self-Reported Fact Check Difficulty} We found that crowdworkers and volunteer fact checkers were not accurate at assessing the difficulty of a fact check, and their assessments of difficulty did not correspond well to accuracy. We ask each fact checker to report the perceived difficulty of the process, either \emph{easy, medium, hard} before they submitted their verdict. We found that their self-reported perceived difficulty did not correlate with their accuracy --- even if evaluators felt the claim was \emph{easy}, they were only 4\% more accurate in accurately checking it. For \emph{medium} and \emph{hard} claims, the accuracy of fact checking was the same. 

\paragraph{Time Taken to Factcheck}

\begin{figure}[t]
    \centering
    \includegraphics[width=\linewidth]{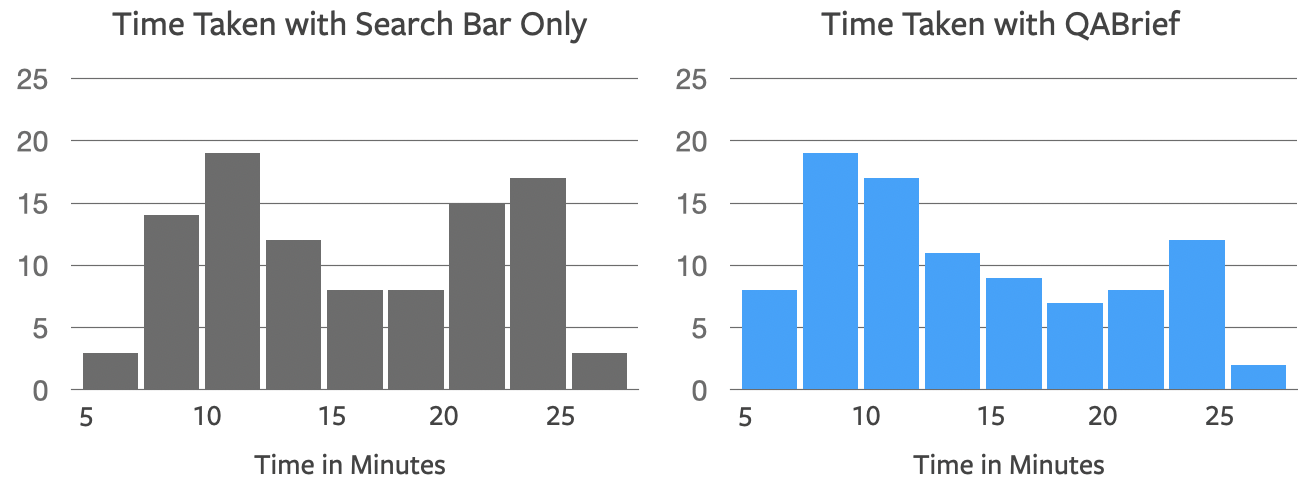}
    \caption{\textbf{Time Taken} to fact check by volunteer evaluators. The distribution is bimodal.}
    \label{fig:time_distribution}
\end{figure}

We present additional results for volunteer fact checking and the time taken to examine claims. As shown in Figure~\ref{fig:time_distribution}, the time taken to fact check is bimodal, most likely because certain claims are easier and others require detailed investigation. Easier claims that were submitted more quickly tended to be checks based more on common sense, for example to fact check the claim \emph{Shark found swimming on freeway in Houston}. When given QABriefs, the distribution of time taken shifts to smaller quantities.

\subsection{Model Training Details}

In this section, we provide detailed information about the training procedure of \qabriefmodel{} as well as exact training and generation parameters used in \texttt{fairseq-py} to produce our results. 

\paragraph{Question Generation}

We use the open sourced BART-large model. We finetune with learning rate $3e-05$, maximum tokens $2048$ per batch, warming up for $500$ updates and training for $10$ epochs. Models are trained with label smoothing $0.1$ and dropout $0.1$. For optimization, we use the Adam optimizer and train with weight decay $0.01$. We tuned only the dropout parameter, between values $0.1, 0.2, 0.3$, but otherwise took these parameters from the suggested parameter settings for BART finetuning. After training, we choose the best checkpoint by validation loss. The total training time is 8 hours on 1 GPU, though reasonable performance is reached after about 5 hours of training. As our model is finetuned BART large, it retains the same parameter count of 406M parameters. 

For generation, we generate with beam size $5$. We tune the length penalty between $0.5, 1, 1.5, 2$ and adjust the minimum and maximum length parameters. For minimum length, we examined values between $3, 5, 10$ and for maximum length, we examined values between $20, 30, 40, 50, 60$. To select the best generation hyperparameters, we generated on the validation set and chose the hyperparameters that maximized BLEU on validation to use on the test set.

\paragraph{Question Answering}

We use the open sourced BART-large model. We finetune with learning rate $3e-05$, maximum tokens $2048$ per batch, warming up for $500$ updates and training for $10$ epochs. Models are trained with label smoothing $0.1$ and dropout $0.1$. For optimization, we use the Adam optimizer and train with weight decay $0.01$. We use the suggested parameter settings for BART finetuning. After training, we choose the best checkpoint by validation loss. The total training time is 8 hours on 1 GPU, though reasonable performance is reached after about 7 hours of training. As our model is finetuned BART large, it retains the same parameter count of 406M parameters.

For generation, we generate with beam size $5$, tuning the beam size between $4, 5$. We keep the length penalty fixed to $1$. We adjust the minimum length parameter between $10, 50$. We adjust the maximum length parameter between $50, 100, 250$. To select the best generation hyperparameters, we generated on the validation set and chose the hyperparameters that maximized BLEU on validation to use on the test set.

\subsection{Dataset Collection Details}

\begin{figure*}[t]
    \centering
    \includegraphics[width=\linewidth]{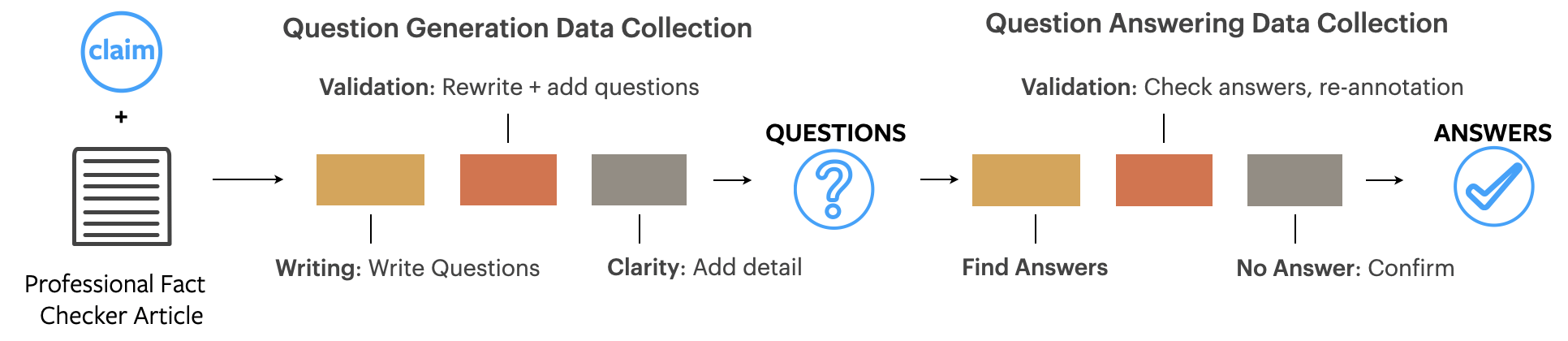}
    \caption{\textbf{\dataset{}}: Annotators read the claim and professionally written fact checking article describing how the claim was checked. Questions and answers are annotated and validated for quality control. Questions are edited so each question is standalone, and \emph{No Answer} options are verified.}
    \label{fig:filtering}
\end{figure*}

In this section, we provide additional details on the instructions given to crowdworkers when constructing \dataset{} and describe all steps. Figure~\ref{fig:filtering} illustrates the full dataset collection process.

\subsubsection{Recruitment for the Task}

We used the crowdworking platform Amazon Mechanical Turk. Evaluators were provided with the task and instructions, and could look at the task and opt to decline. For volunteer fact checkers, volunteers were given a description of our goals and the task they would perform. Volunteers were asked if they were interested in fact checking.

\subsubsection{Question Generation Data Collection}

\paragraph{Instructions for Writing Questions:}
Our goal is to understand how a claim is fact checked. Perform the following steps:

\noindent \fbox{%
    \parbox{\columnwidth}{%
        \begin{itemize}[noitemsep,topsep=0pt] \small
        \item \emph{Read} the claim and the article that describes the fact checking process from a professional fact checker.
        \item \emph{Think} what questions the fact checker had to answer to reach a verdict for the claim
        \item \emph{Write} 3-5 questions that reflect the fact checking process used to reach a verdict. 
        \item Questions must be \emph{standalone} --- do not write questions that refer to other questions, specify the names of people/places, etc 
    \end{itemize}
    }%
}%

\vspace{0.5cm}

\noindent \fbox{%
    \parbox{\columnwidth}{%
    \textbf{Important!}
    \begin{itemize}[noitemsep,topsep=0pt] \small
        \item \emph{DO NOT} write questions with yes or no answers
        \item \emph{DO NOT} write questions that rephrase the claim
    \end{itemize}
    }%
}
\vspace{.5cm}

\noindent \fbox{%
    \parbox{\columnwidth}{%
    \textbf{Must Read Examples:}

    \begin{itemize}[noitemsep,topsep=0pt] \small
        \item \emph{Good:} What was the population of California in 2000?
        \item \emph{Bad:} What was the population of California? [No time specified to find a statistic]
        \\ 
        \item \emph{Good:} How many education bills did Senator Smith vote for in March, 2000?
        \item \emph{Bad:} How many education bills did he vote for? [Who is he? Also no time specified]
        \\ 
        \item \emph{Good:} How do sharks move around?
        \item \emph{Bad:} Is it true that sharks can walk on land? [Yes or no question, and directly asks if something is true or not]
    \end{itemize}
    }%
}
\vspace{0.5cm}

In this data collection step, we used a number of automatic checks implemented into the task. Annotators could not submit without filling out at least 3 questions, each of at least 5 tokens in length. The questions could not overlap with each other more than 5 words. The questions could not exactly match the claim. Annotators could not submit in the first minute of the task. For each problem detected by the automatic check, an error message was displayed explaining why the current submission was not valid. 

\paragraph{Instructions for Validating Questions}: Our goal is to understand the steps necessary to fact check a claim. Perform the following steps:

\noindent \fbox{%
    \parbox{\columnwidth}{%
        \begin{itemize}[noitemsep,topsep=0pt] \small
            \item \emph{Read} the claim and the article that describes the fact checking process
            \item \emph{Read} the questions that describe the steps taken by the fact checker to reach a verdict 
            \item \emph{Write} additional questions \textbf{or} \emph{Choose} no additional questions needed
        \end{itemize}
    }%
}

\vspace{0.5cm}

Additional question writing guidelines are the same as for the original writing questions step. Annotators that write more questions are paid a bonus.

\paragraph{Instructions for Question Clarity}: Our goal is to make sure each question is readable and could be used in a Google search to find an answer. Perform the following steps:

\noindent \fbox{%
    \parbox{\columnwidth}{%
        \begin{itemize}[noitemsep,topsep=0pt] \small
            \item \emph{Read} the question 
            \item Do you think the question could be Googled to find an answer? If not, \emph{read} the article and add more detail to the question 
        \end{itemize}
    }%
}

\vspace{0.5cm}

\noindent \fbox{%
    \parbox{\columnwidth}{%
        \textbf{Must Read Examples:}
        \begin{itemize}[noitemsep,topsep=0pt] \small
            \item \emph{Original:} What was the population of California?
            \item \emph{Edit:} What was the population of California in 2000? [Adds year]
            \\ 
            \item \emph{Original:} How many education bills did he vote for? 
            \item \emph{Edit:} How many education bills did Senator Smith vote for in March, 2000? [Adds name and year]
        \end{itemize}
    }%
}

\subsubsection{Question Answering Data Collection}

\paragraph{Instructions for Finding Answers:} Our goal is to find answers to each of these questions. Perform the following steps:

\noindent \fbox{%
    \parbox{\columnwidth}{%
        \begin{itemize}[noitemsep,topsep=0pt] \small
            \item \emph{Read} the question 
            \item Use the \emph{Search Bar} to find an answer
            \item If you cannot find an answer, you must write an explanation why you cannot find the answer
        \end{itemize}
    }%
}

\vspace{0.5cm}

\noindent \fbox{%
    \parbox{\columnwidth}{%
        \textbf{Important!}
        
        \begin{itemize}[noitemsep,topsep=0pt] \small
            \item \emph{DO NOT} use any other search bar to find an answer. You MUST use the provided search bar only. 
            \item \emph{Do NOT} answer the claim or predict a verdict. Your job is to find an answer to the \emph{QUESTION}
            \item \emph{DO NOT} submit answers from politifact.com or factcheck.org. These answers will not be accepted. If you use our provided search bar, you will not have this problem. Use the provided search bar!
        \end{itemize}
    }%
}

\vspace{0.5cm}

The task then has a dynamic workflow, which we now describe. After using the search bar, annotators had to select between one of three options:

\vspace{0.5cm}

\noindent \fbox{%
    \parbox{\columnwidth}{%
        \begin{itemize}[noitemsep,topsep=0pt] \small
            \item I found an answer, and I can copy paste the text of the answer from the webpage
            \item I cannot copy paste the answer because it is in a graph, table, or picture, but I can write the answer myself.
            \item I cannot find an answer. I understand I will need to write an explanation why an answer cannot be found
        \end{itemize}
    }%
}

\vspace{0.5cm}

and these options correspond to extractive, abstractive, and no answer possibilities. 

If the annotator chose the first option, an extractive answer, they were presented with a form with the following instructions: copy-paste the answer text, copy-paste the URL the answer is from. They are asked to \emph{Copy paste the answer. DO NOT copy paste the entire site, only the part that answers the question. You can paste a maximum of 250 words}. 

If the annotator chose the second option, an abstractive answer, they were presented with a form with the following instructions: write the answer text using at least 20 words, copy-paste the URL the answer is from.

If the annotator chose the third option, no answer, they were presented with a form with the following instructions: write an explanation for why no answer can be found using at least 20 words. 

In this data collection step, we used a number of automatic checks implemented into the task. Annotators could not submit the task unless all requested areas (based on their chosen branch of the workflow) were filled out. The extractive answer could not be more than 250 words in length and could not be the empty string (one word answers were accepted). The abstractive answer and no answer explanation had to be at least 20 words in length. The copy pasted URL the annotators submitted as evidence for their answer had to match the URLs of their returned search results. This serves a dual purpose check --- first that annotators used our search bar, which is restricted from accessing fact checking domains, and second that annotators submitted a real URL. Annotators could not submit in the first minute of the task. Annotators could not submit URLs that were known fact checking domains. For each problem detected by the automatic check, an error message was displayed explaining why the current submission was not valid.

\end{document}

%% file: intro.tex
\begin{figure}[t]
    \centering
    \includegraphics[width=\linewidth]{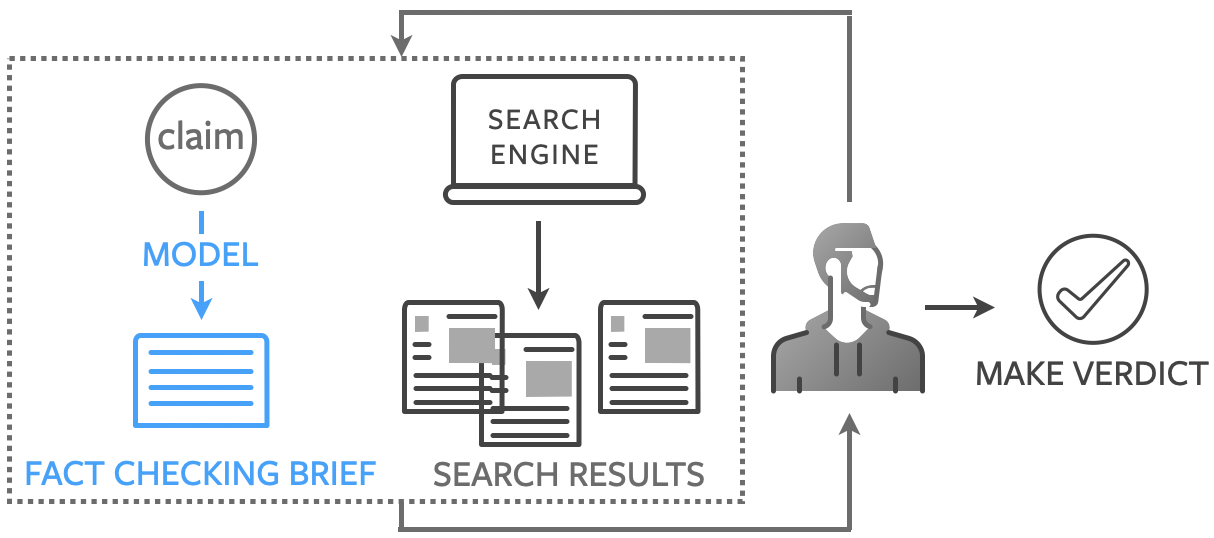}
    \caption{\textbf{Fact Checking Briefs.} Before conducting a fact check, we propose generating \emph{briefs} to provide information about the claim.
    We show they make fact checking more accurate and efficient.}
    \label{fig:motivation}
\end{figure}

Fact checking is a challenging task. It requires deep knowledge of a claim's topic and domain, as well as an understanding of the intricacies of misinformation itself. 
Checking a single claim can take professional fact checkers 15 minutes to one day~\cite{hassan15quest}.
Volunteers on the other hand are not considered accurate enough;

with access to a search engine, \citet{roitero2020can} report crowdsourced fact check accuracies of around 58\%.
This result corroborates earlier reports
\footnote{\url{http://mediashift.org/2010/11/crowdsourced-fact-checking-what-we-learned-from-truthsquad320/},\\ \url{http://fullfact.org/blog/2018/may/crowdsourced-factchecking/}} 
by fact checking websites which attempted to engage volunteers, but reported success only for claim detection, which is considered a much simpler task \cite{konstantinovskiy2018automated}.
This is problematic, both from the perspective of using crowdsourced fact checking to combat misinformation and from the perspective of helping individuals fact check themselves.

\begin{figure*}[t]
    \centering
    \includegraphics[width=\linewidth]{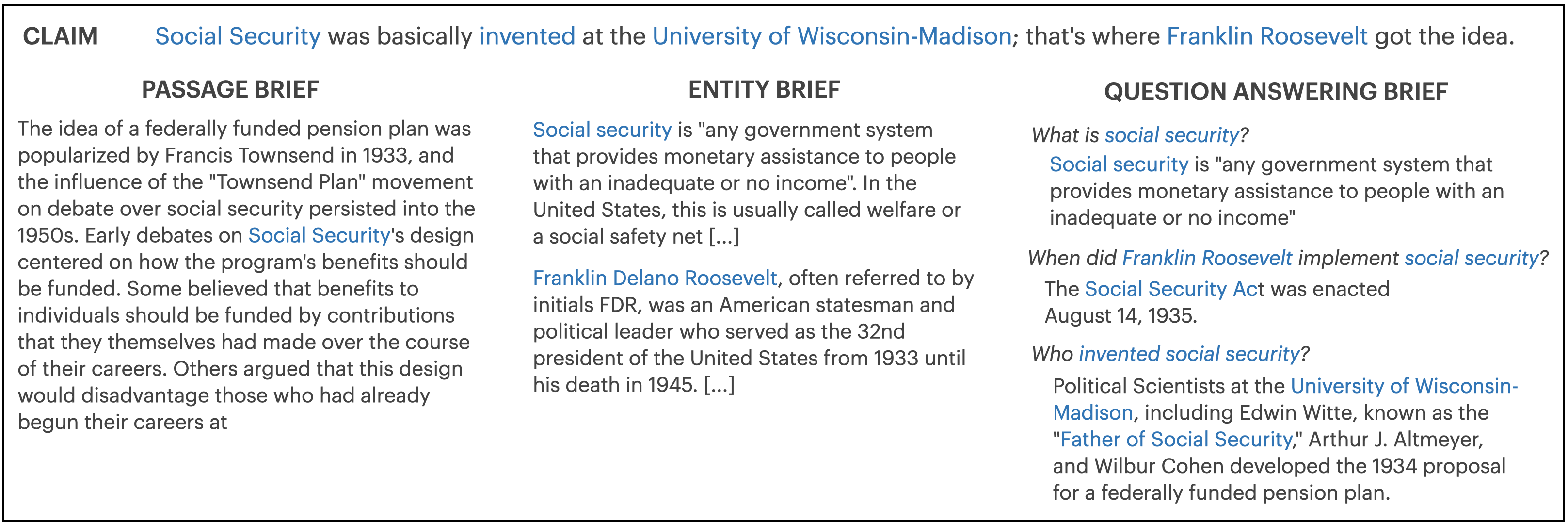}
    \caption{\textbf{Three Types of Briefs:} \textbf{(1) Passage Briefs}, based on information retrieval applied to the claim, \textbf{(2) Entity Briefs}, using entity linking to identify information about each entity, and \textbf{(3) Question Answering Briefs}, which condition on the claim to generate questions, then answer questions using open domain question answering}
    \label{fig:example_briefs}
\end{figure*}

One path for scaling fact checking could be through full automation, taking a claim as input and producing a  verdict~\cite{vlachos-riedel-2014-fact}. Existing work has framed fact checking as classification, often supported by evidence~\cite{wang-2017-liar,thorne-etal-2018-fever,augenstein-etal-2019-multifc}.
However, due to the limitations of existing automated solutions, practitioners prefer solutions that improve efficiency in reaching a verdict, instead of approaches to the complete process \cite{graves2018understanding}.

In this work, we propose \emph{briefs} to increase the accuracy and efficiency of fact checking ( Figure~\ref{fig:motivation}). By generating fact checking briefs, our models aim to provide evidence a human fact checker would find useful.
We investigate several approaches, 
including 
returning Wikipedia passages that relate to the claim, and an entity linking approach that shows information about mentioned entities. Crucially, we  introduce QABriefs --- a set of relevant questions and their answers (see Figure~\ref{fig:example_briefs}).

To learn how to produce QABriefs and create training data, we use  crowdsourcing to gather such briefs based on existing fact checks. We create \dataset{}, a collection of about 10,000 QABriefs with roughly 3 question and answer pairs each. We introduce \qabriefmodel{}, a novel model that performs structured generation via claim-conditioned question generation and open domain question answering. 
Each question is used to identify evidence using a search engine. Finally, a pretrained question answering model is finetuned to generate answers and produce the full brief.  

In experiments with crowdworkers, QABriefs improve accuracy by 10\% compared to using only a search bar while reducing the time a fact check takes.  
For volunteer 
fact checkers, accuracy is improved by 4\% and the process is 20\% faster compared to using a search bar. 
Using QABriefs from human annotators leads to the largest improvement, followed by briefs generated by \qabriefmodel{} and  other proposed forms of briefs. 
This suggests that briefs are a promising avenue for improving crowdsourced fact checking. 
Further, \dataset{} can be used to develop models capable of answering challenging, real world questions.